\documentclass[sigconf]{acmart}
\usepackage{latexsym}
\usepackage{listings}
\usepackage{subcaption}
\usepackage{enumitem}
\usepackage{algorithm}
\usepackage{graphicx}
\usepackage[noend]{algorithmic}
\usepackage{tabularx}
\usepackage{multirow}
\usepackage{makecell}
\usepackage{parcolumns}
\usepackage[normalem]{ulem}

\usepackage{bm}

\usepackage[most]{tcolorbox}
\usepackage{xcolor}

\usepackage{booktabs}
\usepackage{etoolbox}
\makeatletter
\definecolor{yellow}{RGB}{255,255,168}
\definecolor{lightblue}{RGB}{0, 255, 255}
\definecolor{green}{RGB}{64, 224, 208}
\definecolor{orange}{RGB}{255, 195, 0}
\definecolor{gray}{RGB}{220,220,220}
\definecolor{lightgreen}{RGB}{144, 238, 144}
\definecolor{pink}{RGB}{255, 192, 203}

\AtBeginDocument{%
  \providecommand\BibTeX{{%
    \normalfont B\kern-0.5em{\scshape i\kern-0.25em b}\kern-0.8em\TeX}}}

\setcopyright{acmcopyright}
\copyrightyear{2018}
\acmYear{2018}
\acmDOI{XXXXXXX.XXXXXXX}

\acmConference[Conference acronym 'XX]{Make sure to enter the correct
  conference title from your rights confirmation emai}{June 03--05,
  2018}{Woodstock, NY}
\acmPrice{15.00}
\acmISBN{978-1-4503-XXXX-X/18/06}

\begin{document}

\title{LAMP: A Language Model on the Map}


\author{Pasquale Balsebre}
\affiliation{
    \institution{Nanyang Technological University}
   \country{Singapore}
 }
 \email{pasquale001@e.ntu.edu.sg}

\author{Weiming Huang}
 \affiliation{
   \institution{Nanyang Technological University}
   \country{Singapore}}
 \email{weiming.huang@ntu.edu.sg}

 \author{Gao Cong}
 \affiliation{
   \institution{Nanyang Technological University}
   \country{Singapore}}
 \email{gaocong@ntu.edu.sg}



\begin{abstract}
Large Language Models (LLMs) are poised to play an increasingly important role in our lives, providing assistance across a wide array of tasks. In the geospatial domain, LLMs have demonstrated the ability to answer generic questions, such as identifying a country's capital or recommending touristic attractions; nonetheless, their utility is reduced when it comes to answering fine-grained questions about specific places, such as grocery stores or restaurants, which constitute an essential aspect of people's everyday life. This is mainly because such places haven't been systematically fed into LLMs, so as to learn about them. This study introduces a novel framework for fine-tuning a pre-trained model on city-specific data, to enable it to provide accurate geospatial recommendations in a conversational fashion, while minimizing hallucinations. We propose LAMP, and conduct experiments to analyze its ability to correctly retrieving spatial objects, and compare it to well-known open- and closed- source language models, such as GPT-4. Finally, we explore its emerging capabilities through a case study on day planning.


\end{abstract}

\ccsdesc[500]{Computer systems organization~Embedded systems}
\ccsdesc[300]{Computer systems organization~Redundancy}
\ccsdesc{Computer systems organization~Robotics}
\ccsdesc[100]{Networks~Network reliability}

\keywords{datasets, neural networks, gaze detection, text tagging}



\maketitle

\section{Introduction}
LLM-based virtual assistants, such as ChatGPT, being pre-trained on Internet-scale text corpora, possess general knowledge in various domains, ranging from medicine and history, to math and computer science \cite{llms_few_shot}. Researchers have conducted studies to assess the capabilities of language models also in various geospatial applications, including health geography \cite{llms_for_health}, remote sensing \cite{remote_sensing}, and urban geography \cite{mai2023opportunities}, revealing promising avenues for the development of geospatial artificial intelligence (GeoAI) \cite{mai2023opportunities, cityfm}. 

\begin{figure*}
  \centering
  \includegraphics[width=1\textwidth]{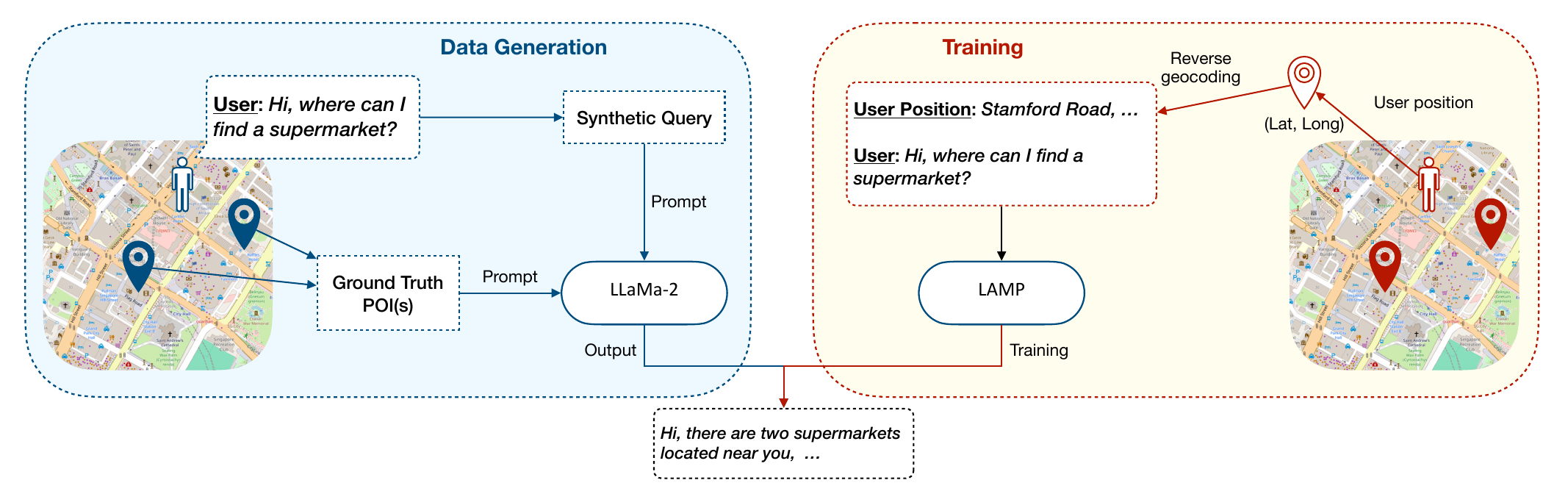}
  \caption{The proposed framework: left hand side shows the Data Generation process, while the right hand side illustrates the strategy adopted to train LAMP.}
  \label{framework_figure}
\end{figure*}

\begin{lstlisting}[caption={Example responses of LLM-based assistants when queried to find a Point of Interest. Orange: hallucinated information. Yellow: inconclusive information. Green: correct information.},captionpos=t, basicstyle=\scriptsize\ttfamily, escapeinside={@}{@}, literate={hrule}{{\hrulefill}}5, label=lst:listing1]
@\colorbox{lightblue}{Current position:}@ 310 Orchard Rd, Singapore 238864
@\colorbox{green}{Query:}@ Can you find a nearby Starbucks?
hrule
@\colorbox{pink}{Claude-2:}@ Sure, [...] there is a @\colorbox{orange}{Starbucks at 390 Orchard}@
@\colorbox{orange}{Road, \#01-02/03, Palais Renaissance, Singapore 238871}@ [...]
@\colorbox{orange}{approximately 0.3 miles or 6 minutes walk from your position}@
hrule
@\colorbox{pink}{Claude-3 Sonnet:}@ Sure, here are a few Starbucks near your location:
1. @\colorbox{lightgreen}{Starbucks (Ngee Ann City). Address: 391 Orchard Road}@, 
@\colorbox{orange}{\#B2-33 Ngee Ann City, Singapore 238872}@
2. @\colorbox{orange}{Starbucks (Orchard Central) Address: 181 Orchard Road, \#01-28/29 Orchard }@ 
@\colorbox{orange}{Central, Singapore 238896}@
3. @\colorbox{orange}{Starbucks (Orchard Paragon). Address:}@ @\colorbox{lightgreen}{290 Orchard Road}@, 
@\colorbox{orange}{\#B1-03/04 The Paragon,}@ @\colorbox{lightgreen}{Singapore 238859}@
hrule
@\colorbox{pink}{ChatGPT:} @  @\colorbox{yellow}{I'm sorry but I don't have the capability to access}@
@\colorbox{yellow}{real-time data, including specific locations of businesses}@ [...]
hrule
@\colorbox{pink}{LAMP:} @  Sure, there is a Starbucks close to your location, its
address is @\colorbox{lightgreen}{2 Orchard Turn, \#B3 - 59, Singapore 238801}@, it is
located @\colorbox{lightgreen}{inside ION Orchard}@ shopping mall, provides @\colorbox{lightgreen}{wheelchair}@
 @\colorbox{lightgreen}{access}@ and it's open until @\colorbox{lightgreen}{10pm}@.
\end{lstlisting}

Despite that, most explorations on LLMs for geospatial applications rely on the engineering of textual prompts, to induce LLMs to provide pertinent results for specific downstream geospatial applications. Popular LLMs, like the GPT-family models, lack detailed knowledge, e.g., location, name, and address, for specific Points of Interest (POIs), because such detailed geospatial information is not well covered in the pre-training corpora. This is justified by the fact that geospatial data and applications are considered particularly specialized.

In this context, existing models, like ChatGPT, can successfully answer generic geography questions such as "\textit{What is the capital of France?}" or "\textit{Which monuments can I visit in Paris?}", whereas they tend to provide unsatisfactory answers, when presented a query that requires knowledge about specific places in a particular area. The query in Listing \ref{lst:listing1} is an example of a location-based POI-search query. As shown, Claude shows hallucinated information about a place that does not exist, whereas ChatGPT admits not to have access to information regarding the locations of businesses. Developing an LLM that is capable of effectively answering specific geospatial questions, particularly for POI retrieval, in a conversational form, has tremendous utility in our daily lives, as finding such places is a prevalent need for most people.


In fact, an LLM with conversational capabilities and knowledge about geospatial entities, can provide assistance in a wide range of tasks, such as urban functionality analysis, tourist trip planning, property search, and healthcare. For example, in case of a medical emergency, such a model can swiftly identify a close-by hospital, that is also suitable for the type of emergency.

This study proposes a novel framework to incorporate geospatial knowledge of a specific area, into a pre-trained LLM, by fine-tuning it on a self-supervised task. We describe a process for automated generation of data through Retrieval-Augmented Generation (RAG) \cite{retrievalaugmentedgen}, and introduce a viable path to mitigate the hallucination problem. During the fine-tuning phase, LAMP not only learns about the existence and locations of geospatial objects, but also grasps the proximity between various streets and districts in the city. This enables LAMP to consistently offer relevant suggestions based on the user's position.

We conduct experiments to analyze LAMP's ability to accurately retrieving spatial objects, assess the relevance and correctness of its responses, and compare it to well-known open- and closed-source LLMs. Finally, we explore the emerging capabilities of LAMP through a case study involving planning a day out and addressing complex natural language queries. In summary, in this paper, we:

\begin{itemize}[leftmargin=*]
  \item Introduce a new paradigm for utilizing LLMs in geospatial applications, by letting the model memorize the spatial objects inside an urban area, and subsequently using it to address conversational queries about such places.
  \item Propose a simple, yet effective, framework to inject geospatial knowledge into a pre-trained LLM. This consists of knowledge about POIs, their locations, and related information. The process guides the model in gaining spatial (proximity) awareness in a specific urban area of interest.
  \item Assess, with the help of a team of GIS-domain experts, the performance of the proposed LAMP and the selected baselines, in terms of truthfulness, spatial awareness and semantic relatedness.
  \item Share the data\footnote{https://github.com/PasqualeTurin/LAMP/tree/main/data} to train LAMP, consisting of 113,000 conversational interactions, and, upon acceptance, the model's parameters.
\end{itemize}

\section{Methodology}
The primary objective of this study is to integrate information about geospatial objects in a given region of interest into an LLM, so as to make it a helpful assistant for geospatial tasks within the selected region. This entails ensuring the model is aware of the existence of each spatial object, and able to grasp proximity between them and the user that will be querying it. To this end, we collected 18,390 Points of Interest from Yelp Singapore, which provides the position and a complete address for each POI. Besides, additional information such as opening hours, price range, and rating are often available. We selected LLaMa-2-7B-Chat \cite{llama2} as the LLM to be fine-tuned on this task, but the same framework can be applied to any other language model.

The workflow to generate the data and to fine-tune LAMP, is depicted in Figure \ref{framework_figure}. The left hand side of the image shows the data generation phase, during which synthetic queries are used to generate conversational interactions about one or more POIs. To ensure that each POI is included in the training corpus, and that the model learns about it, we generate $N_t$ (= 10) queries for each POI or its category, from a position randomly sampled inside a circle of radius $r$ (= 150m) centered on the POI itself. Table \ref{tab:queries} illustrates some examples of the different types of synthetic queries that we generate. The original version of LLaMa-2-7B-Chat is then used to generate a reply, using a prompt that includes structured information about the POI(s) that should be retrieved to satisfy the user's request. 
The right hand side of Figure \ref{framework_figure} shows that LAMP is trained to predict the response generated by LLaMa, while receiving in the prompt only the information regarding the user's position. We use Nominatim, the OSM reverse geocoding API, to translate the position of the user into an address, being the latter more suitable for an LLM. The intuition behind this training strategy, is that the model will be forced to learn that those POIs exist, and that their positions (expressed as addresses) are in spatial proximity with the user's current position.

Empirically, we observed that training the model solely on queries that can always be satisfied with a nearby POI, leads it to replicate the training behavior, irrespective of the actual existence of the POI. To alleviate this issue, we included in the training corpus, queries from random positions in the city, incorporating random place names or categories, thus letting the model learn that nearby POIs are not always available, and that an existing POI located further away is a more suitable alternative to a hallucinated one.

\begin{figure}
  \centering
  \caption{Training and validation loss, during the 5 epochs of training.}
  \includegraphics[width=0.35\textwidth]{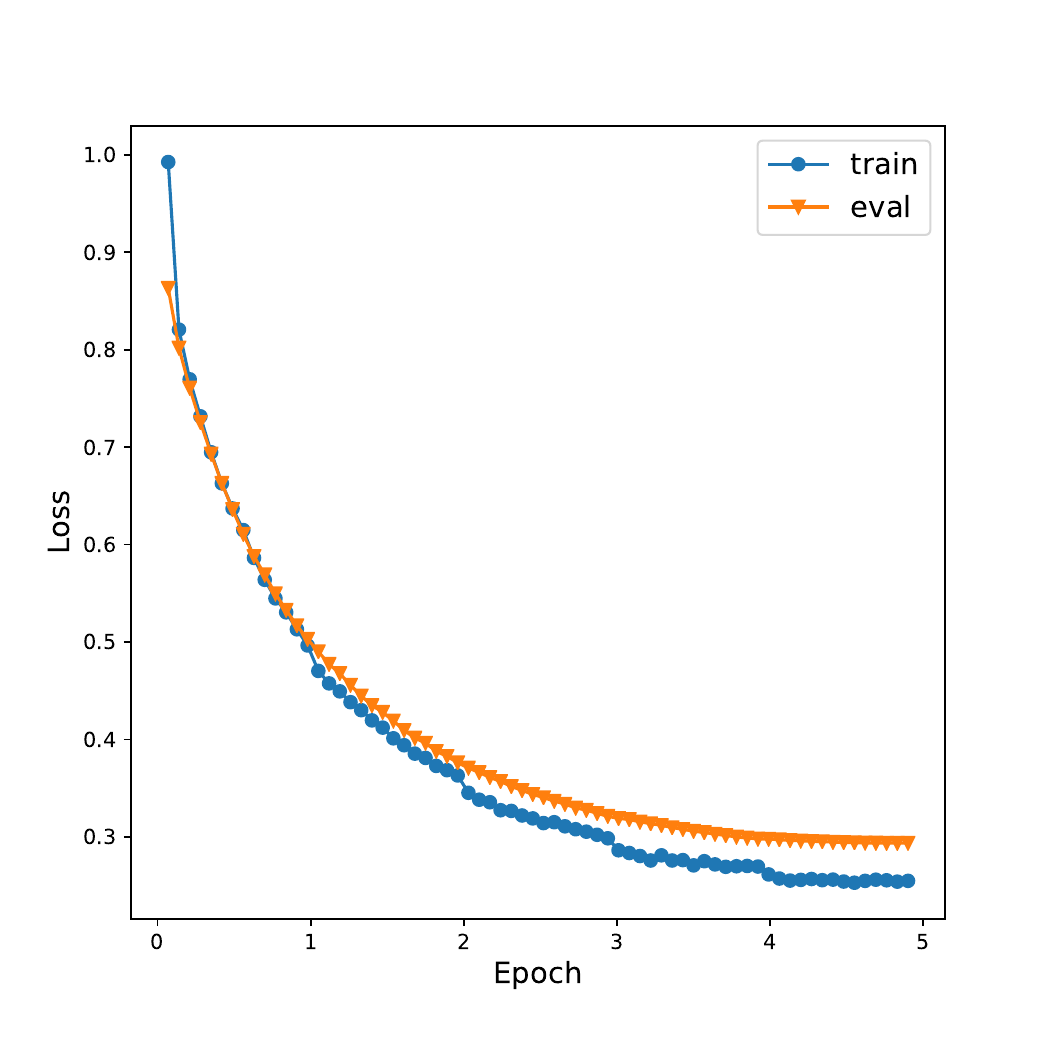}
  \label{loss_figure}
\end{figure}

\begin{table}[ht]
\centering
\caption{Example of the different type of synthetic queries, used as LAMP's training data.}
\label{tab:queries}
\begin{tabular}{lp{5cm}}
\toprule
\textbf{Type} & \textbf{Query example} \\
\midrule
\scalebox{0.95}{Name search} & Hi LAMP, tell me where is [\textit{POI\_Name}] located \\
\hline
\scalebox{0.95}{Category search} & Can you please help me finding a nearby [\textit{POI\_Category}]\\
\hline
\scalebox{0.95}{Type search} & Can you please point out a highly rated restaurant in the area?\\
\hline
\scalebox{0.95}{Type search} & Can you please point out a nearby restaurant that offers [\textit{food\_type}] food?\\
\bottomrule
\end{tabular}
\end{table}

\section{Experiments}
In this section we illustrate the settings used to train and test LAMP, and discuss the main findings that emerged from our experiments.

\subsection{Training Settings}
We generated 6 synthetic queries for each POI, leading to a dataset of  $\sim$110,000 queries. Five queries for each POI were allocated for training, while the remaining one was reserved for validation. LAMP was trained for 5 epochs, starting from the LLaMa-2-7B-Chat weights, on a single A100, 80GB GPU, for 78 hours. We used 4-bit quantization \cite{quantization} to obtain a more compact model representation, and low rank adaptation (LoRA) \cite{lora} to reduce the number of trainable parameters and decrease the GPU memory requirements. Image \ref{loss_figure} shows the training and validation loss, over the 5 epochs. The model was validated every $\sim$0.07 epochs and a model checkpoint was saved. As the figure shows, the training process was smooth, and the model converged towards the fifth epoch.

\subsection{Test Settings}
We tested the performance of our model in a typical POI-search scenario: a user queries the model for suggestions about a place to visit, or a specific desire (e.g., "\emph{I would like to buy new clothes}"). We assumed that the user's position is available, converted it to an address using reverse geocoding, and provided it to the model in the prompt, as in the example in Listing \ref{lst:listing1}. We asked a team of GIS-domain experts, to design 50 queries issued from random positions in Singapore, and evaluate the quality of the suggestions provided by a set of LLM-based assistants. Specifically, the truthfulness (does the POI suggested by the model exist?), spatial awareness (is the POI suggested by the model, the closest?) and semantic relatedness (is the POI suggested by the model related to the request?) of the replies were assessed.

\subsection{Baselines}
We compare LAMP to both open- and closed-source models of different sizes:
\begin{itemize}[leftmargin=*]
  \item \textbf{ChatGPT 3.5}, \textbf{ChatGPT 4o} and \textbf{ChatGPT 4.0} are the main versions of the popular LLM-based AI assistant ChatGPT. Notably, ChatGPT-4o and ChatGPT-4, with \emph{Browsing} activated, have the capability of retrieving results from the web.
  \item \textbf{Claude-2} and \textbf{Claude-3 Sonnet}\footnote{https://claude.ai/} are powerful language models developed by Anthropic. They excel at sophisticated dialogue and creative content generation, as well as detailed instruction provision.
  \item \textbf{LLaMa-2 7B}, \textbf{LLaMa-2 13B} and \textbf{\textbf{LLaMa-2 70B}} are respectively the small, medium and large iterations of the popular Language Model LLaMa-2 \cite{llama2}.
  \item \textbf{LAMP} and \textbf{LAMP-no-rq}. LAMP is our main model and LAMP-no-rq is an identical model, trained excluding queries from random positions about random places or categories. We employ this version to assess the influence of including such queries on the model's hallucinations.
\end{itemize}

\begin{table}
    \centering
    \caption{Main results on POI-retrieval in Singapore.}
    \begin{tabular}{cccc}
         \toprule
         \multirow{2}{*}{Model} & \multirow{2}{*}{Truthfulness} & Spatial & Semantic \\[-0.05cm]
         &&Awareness&Relatedness\\[0.cm]
         \toprule
         \scalebox{0.9}{LLaMa-2-7B} & 0.12 & 0.2 & 0.76\\[0.15cm]
         \scalebox{0.9}{LLaMa-2-13B} & 0.18 & 0.23 & 0.88\\[0.15cm]
         \scalebox{0.9}{LLaMa-2-70B} & 0.3 & 0.36 & 0.94\\
         \midrule
         \scalebox{0.9}{Claude-2} & 0.22 & 0.32 & 0.96\\[0.15cm]
         \scalebox{0.9}{Claude-3 Sonnet} & 0.49 & 0.48 & 0.96\\
         \midrule
         \scalebox{0.9}{ChatGPT-3.5} & 0.68 & 0.6 & \uline{0.98}\\[0.1cm]
         \scalebox{0.9}{ChatGPT-4o} & \multirow{2}{*}{\uline{0.92}} & \multirow{2}{*}{\uline{0.84}} & \multirow{2}{*}{\textbf{1.0}}\\[-0.15cm]
         \scalebox{0.7}{(Browsing)}&&&\\[0.1cm]
         \scalebox{0.9}{ChatGPT-4} & \multirow{2}{*}{\textbf{0.94}} & \multirow{2}{*}{0.82} & \multirow{2}{*}{\textbf{1.0}}\\[-0.15cm]
         \scalebox{0.7}{(Browsing)}&&&\\
         \midrule
         \scalebox{0.9}{LAMP-no-rq} & 0.76 & \uline{0.84} & \textbf{1.0}\\[0.15cm]
         \scalebox{0.9}{LAMP} & 0.86 & \textbf{0.92} & \textbf{1.0}\\
         \bottomrule
    \end{tabular}
    \label{tab:results}
\end{table}

\subsection{Findings}
The main results of the experiments on conversational POI-retrieval are reported in Table \ref{tab:results}.
\subsubsection{Truthfulness}
In this subsection, we assess the capability of baseline models to provide information on existing POIs and related factual details. LAMP's truthfulness score of 86\%, showcased an increase of 10\% compared to LAMP-no-rq, in absolute terms. In our experiments we observed that when a nearby POI is unavailable or the model fails to retrieve it, LAMP-no-rq tends to provide a non-existent POI, locating it near the user, while LAMP retrieves a real POI, albeit sub-optimal in terms of distance. This is because LAMP-no-rq is trained on queries for which an ideal POI to be retrieved, is always available at a short distance, thus teaching the model that this is always the case.

ChatGPT-4o and ChatGPT-4 showcase the highest truthfulness scores of 92\% and 94\%. Although the replies are based on the browsing results, we observed that the models sometimes produce hallucinated or outdated results. ChatGPT-3.5 and the LLaMa-2 models exhibited lower truthfulness, likely due to the lower model size and coverage in the training set.

\subsubsection{Spatial awareness}
In this subsection, we analyze the spatial awareness of the models by assessing the frequency with which they recommend POIs located in close proximity to the user. LAMP excelled particularly in spatial awareness, with an accuracy of 92\%. Thanks to the extensive training on queries from thousands of positions, LAMP learned a model of Singapore, gaining a granular knowledge about the POIs, roads and districts of the city, as well as spatial proximity among them. ChatGPT-4o achieved a spatial awareness score of 84\%. ChatGPT 3.5, Claude and the LLaMa-2 models base their answers on their internal weights, without performing any web search, delivering much less useful and reliable results. We noticed that ChatGPT 3.5 can often retrieve POIs that exist, thanks to the extensive coverage of its training data. However, it lacks information about more specific places and spatial knowledge, and in most cases provides places that are not in spatial proximity to the user, or very generic suggestions.

\subsubsection{Semantic relatedness}
All the models analyzed provided replies that were relevant with the user query, in the majority of the cases. ChatGPT-4 and our LAMP models attained a perfect score, albeit ChatGPT-4 often required query reformulation to provide specific Points of Interest (POIs), to force it not to offer generic suggestions of city regions that might interest the user.

\begin{figure}
  \centering
  \includegraphics[width=0.4\textwidth]{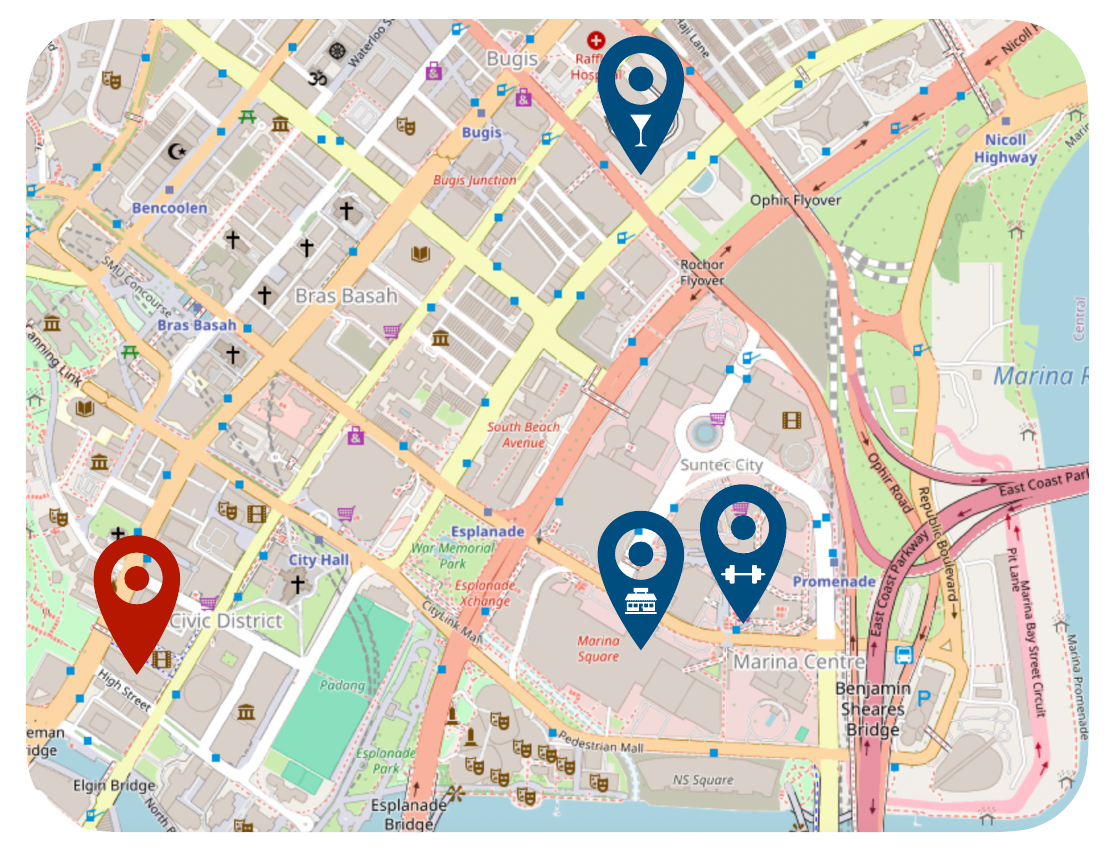}
  \caption{The POIs recommended by LAMP, in response to the query in Listing \ref{lst:planning}.}
  \label{casestudy_figure}
\end{figure}

\begin{lstlisting}[caption={Day out planning},captionpos=t, basicstyle=\scriptsize\ttfamily, escapeinside={@}{@}, literate={hrule}{{\hrulefill}}5, label=lst:planning]
@\colorbox{lightblue}{Current position:}@ 100 High St, Singapore 179434
@\colorbox{green}{Query:}@ Hi LAMP, we are planning a day out. We would like to find
a nearby gym to exercise, then buy something at a convenience
store, and finally head to a bar for some drinks. Could you
provide some recommendations?
hrule
@\colorbox{pink}{ChatGPT-4o:} @ Sure, I can help you plan your day out. Let's find a
nearby gym, a convenience store, and a bar around your position.

@\colorbox{lightgreen}{Pure Fitness - Asia Square Tower 1}@
Address: @\colorbox{lightgreen}{8 Marina View}@, @\colorbox{orange}{\#02-03}@, @\colorbox{lightgreen}{Asia Square Tower 1, 018960}@
Features: Group classes, personal training.

@\colorbox{orange}{Cheers - Funan Mall}@
@\colorbox{orange}{Address: 107 North Bridge Road, \#B2-14, Funan Mall}@
Features: Extensive selection of groceries.

@\colorbox{yellow}{The Other Room}@
@\colorbox{yellow}{Address: 320 Orchard Rd, \#01-05 Marriott Tang Plaza Hotel, 238865}@
Features: A stylish bar known for its sophisticated atmosphere.
hrule
@\colorbox{pink}{LAMP:} @ Sure! Here are the places that I recommend:
1. First, visit True Fitness, at 9 Raffles Boulevard, 039596.
2. Then, proceed to 7-Eleven, located at the first floor of
Marina Square, at 6 Raffles Boulevard, 039594. It offers 
grab-and-go bites and it is open 24 hours.
3. Finally, you can check Bar Square, located at 5 Fraser St,
Singapore 189352. It closes at 10:30pm, offers wheelchair-
accessible entrance, free Wi-Fi and a cosy/casual atmosphere.
Please let me know if you would like me to recommend more places.
\end{lstlisting}

\section{Handling Complex Queries}
LAMP possesses detailed knowledge about Singapore and its Points of Interest. It can correctly retrieve POIs, without accessing any external knowledge base. Because of that, LAMP can be used to address complex queries about places in a conversational fashion. 

For instance, the query in Listing \ref{lst:planning} involves searching for multiple places and creating a short plan. This type of conversational query cannot be addressed by a traditional keyword-based search algorithm, nor by a general-purpose LLM-based assistant, which lacks knowledge of specific places. Although several possible plans exist, and identifying the ideal one can be hard since it depends on the user's needs, we manually verified the plans proposed by GPT-4o and LAMP.
Despite its ability to browse the web, GPT-4o creates an unrealistic plan, in response to our query. As reported in Listing \ref{lst:planning}, it suggests an existing nearby place (light green), \textit{Pure Fitness}, followed by a non-existent convenience store (orange), \textit{Cheers - Funan Mall}, and a bar located at 4.9kms away from the user (yellow), \textit{The Other Room}.
Conversely, all the POIs recommended by LAMP exist, their address and services are correct, and they are located very close to each other. As shown in Figure \ref{casestudy_figure}, the POIs suggested by LAMP are within walking distance to the user's initial position.



\bibliographystyle{ACM-Reference-Format}
\bibliography{sample-base}

\end{document}